\documentclass[10pt,twocolumn,letterpaper]{article}

\usepackage{wacv}
\usepackage{times}
\usepackage{epsfig}
\usepackage{graphicx}
\usepackage{amsmath}
\usepackage{amssymb}
\usepackage{ifthen} %set boolean
\usepackage{color}  %set color
\usepackage{multirow}
\usepackage{paralist}
\graphicspath{{figures/}}
\usepackage{caption}
\usepackage{booktabs}

\usepackage{xspace}
\usepackage{color}
\usepackage{adjustbox}
\usepackage{subfigure}
\usepackage{enumitem} % adjust enum margin
\usepackage{array}
\usepackage{mathrsfs}

\usepackage{comment}

\newcommand{\datasetname}{\text{360-Indoor}\xspace}
\newcommand{\threesix}{\text{360$^\circ$}\xspace}

\newcommand{\Paragraph}[1]{{\flushleft{\textbf{#1}}}}
\long\def\ignorethis#1{}

\usepackage[pagebackref=true,breaklinks=true,letterpaper=true,colorlinks,bookmarks=false]{hyperref}
\newcommand\blfootnote[1]{%
  \begingroup
  \renewcommand\thefootnote{}\footnote{#1}%
  \addtocounter{footnote}{-1}%
  \endgroup
}

\wacvfinalcopy % *** Uncomment this line for the final submission

\def\wacvPaperID{****} % *** Enter the wacv Paper ID here

\ifwacvfinal\fi
\title{\datasetname: Towards Learning Real-World Objects in \threesix Indoor Equirectangular Images}

\author{Shih-Han Chou$^{1}$, Cheng Sun$^{1}$, Wen-Yen Chang$^{1}$, Wan-Ting Hsu$^{1,*}$, Min Sun$^{1}$, Jianlong Fu$^{2}$\\$^{1}$National Tsing Hua University, Hsinchu   $^{2}$Microsoft Research, Beijing\\ {\tt\small shchou75@gmail.com, chengsun@gapp.nthu.edu.tw, $\{$s0936100879, cindyemail0720$\}$@gmail.com}, \\{\tt\small sunmin@ee.nthu.edu.tw, jianf@microsoft.com}}

\ifwacvfinal\fi
\begin{document}
\twocolumn[{%
\renewcommand\twocolumn[1][]{#1}%
\maketitle
\begin{center}
    \captionsetup{type=figure}
	\footnotesize
	\vspace{-3em}
	\includegraphics[width=.95\linewidth]{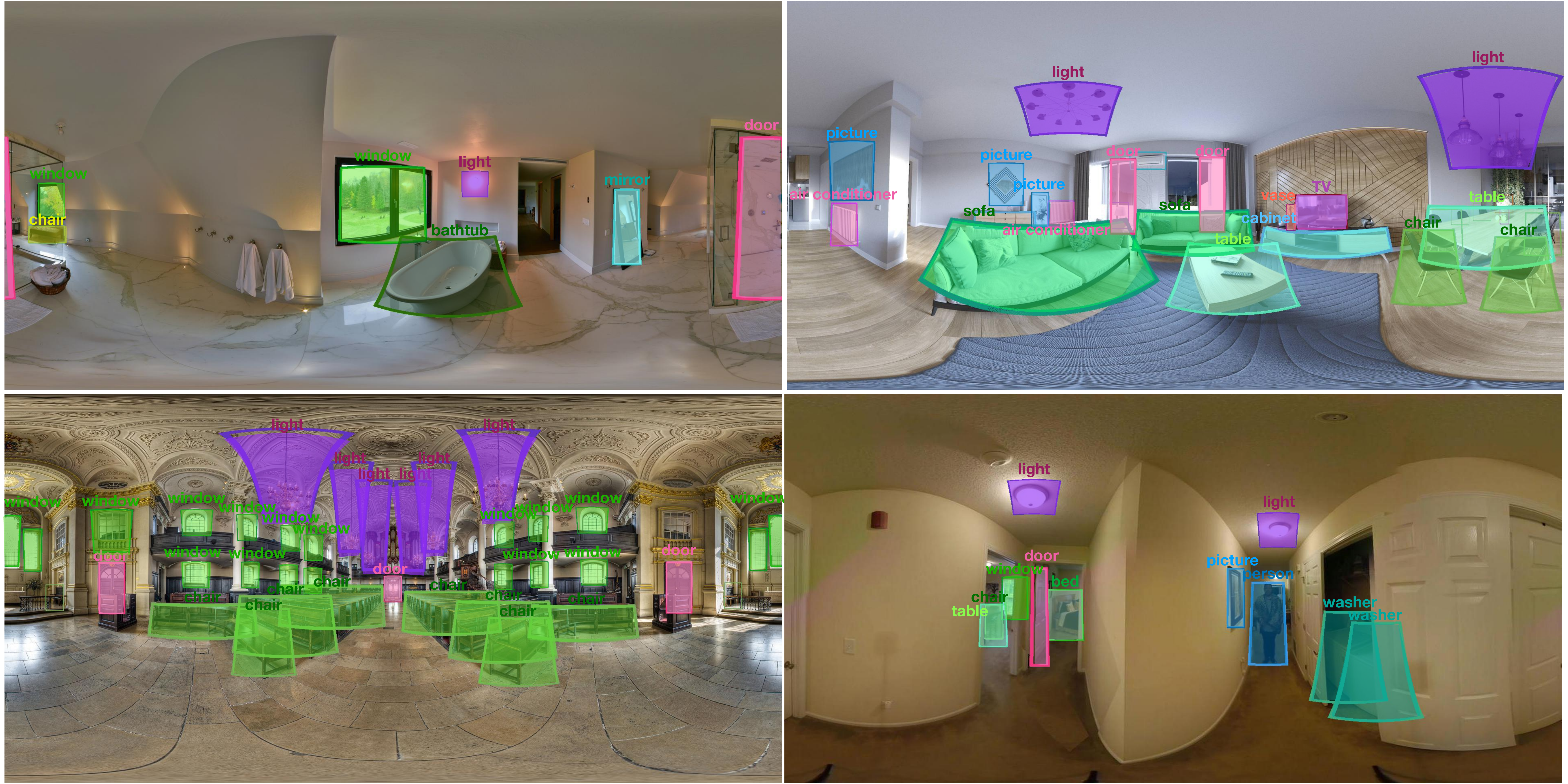}
	\vspace{-1em}
	\captionof{figure}{
	    Examples of the images overlaid with labeled Bounding Field-of-Views (BFoVs) as well as object categories (text and color-code) in our 360-Indoor dataset.
	}
	\label{fig:problem}
\end{center}}]

\blfootnote{*This work was performed when Wan-Ting Hsu was visiting Microsoft Research as a research intern.}
%%%%%%%%% ABSTRACT
\begin{abstract}
While there are several widely used object detection datasets, current computer vision algorithms are still limited in conventional images. Such images narrow our vision in a restricted region. On the other hand, \threesix images provide a thorough sight. 
In this paper, our goal is to provide a standard dataset to facilitate the vision and machine learning communities in \threesix domain. To facilitate the research, we present a real-world \threesix panoramic object detection dataset, \datasetname, which is a new benchmark for visual object detection and class recognition in \threesix indoor images. It is achieved by gathering images of complex indoor scenes containing common objects and the intensive annotated bounding field-of-view. In addition, \datasetname has several distinct properties: (1) the largest category number ($37$ labels in total). (2) the most complete annotations on average ($27$ bounding boxes per image).
The selected $37$ objects are all common in indoor scene. With around $3$k images and $90$k labels in total, \datasetname achieves the largest dataset for detection in \threesix images.
In the end, extensive experiments on the state-of-the-art methods for both classification and detection are provided. We will release this dataset in the near future.
\end{abstract}

%%%%%%%%% BODY TEXT
%%%%%%%%%%%%%%%%%%%%%%%%%%%%%%%%%%%%%%%%%%%%%%%%
%%  Introduction
%%%%%%%%%%%%%%%%%%%%%%%%%%%%%%%%%%%%%%%%%%%%%%%%
\vspace{-1.5em}
\section{Introduction}\label{intro}
Object detection is an essential task in computer vision. The widely used datasets such as MSCOCO~\cite{lin2014microsoft}, Pascal VOC~\cite{Everingham15} have endorsed current researches make huge breakthroughs on object detection tasks~\cite{ren2015faster, long2015fully, wei_2016_SSD, he2017mask, redmon2016you, joseph_2017_yolo9000, redmon2018yolov3}. 
\begin{table*}
    \centering
    \footnotesize
    \caption{Existing \threesix dataset comparison in 2D domain. We list the released dataset to-date.}
    \vspace{-1em}
    \label{tab:datasecompare}
    \begin{tabular}{c|cccccccc}
        \toprule
        \textbf{Dataset} & \textbf{Type} & \textbf{Domain} & \textbf{Purpose} & \textbf{Annotation} & \textbf{\#Category} & \textbf{\#Boxes}\\
        \midrule
        Pano2Vid~\cite{su2016activity}      & Video & Outdoor Activities & Automatic Cinematography & - & - & - \\
        Sports-360~\cite{HuCVPR17}    & Video & Sports & Visual Pilot & Viewing Angles \\
        YouTube/Vimeo~\cite{yu2018deep} & Video & Wedding/Music & Highlight Detection & - & - & - \\
        Narrated-360~\cite{chou2018self}  & Video & House/Tour Guiding & Visual Grounding  & Bounding Boxes & - & -  \\
        Wild-360~\cite{cheng2018cube}      & Video & Nature/Wildlife & Saliency Detection & Saliency Map & - & -  \\
        SUN360\footnotemark~\cite{xiao2012recognizing} & Image & Indoor/Outdoor & Scene/viewpoint recognition  &  Place Categories/Viewpoints & 80 & - \\
        ERA~\cite{yang2018object} & Image & Dynamic Activities & Object Detection & Bounding FoVs & 10 & 7,199 \\
        FlyingCars~\cite{coors2018spherenet} & Image & Synthesis Cars & Object Detection & Bounding FoVs & 1 & 6,000  \\
        \midrule
        \textbf{\datasetname} & Image & Indoor Objects & Object Detection & Bounding FoVs & \bf{37} & \bf{89,148}  \\
        \bottomrule
    \end{tabular}
\end{table*}
Recently, \threesix cameras become more popular and closer to our life because of the wide field of view and the applications to robots and virtual reality~\cite{HuCVPR17, su2017making}. Enormous \threesix videos, such as house guiding, sports, are becoming viral on YouTube. With the growing amount of data, there is an increasing interest in computer vision to dig into \threesix visual recognition. Among numerous \threesix projection (i.e., cubemaps, equirectangular and equiangular cubemaps), the most popular representation of \threesix images is the equirectangular projection. It maps the latitude and longitude of the spherical to horizontal and vertical grid coordinates.

However, significant distortions of equirectangular images is a crucial problem, especially in the polar regions.
Although many works propose spherical convolutional neural networks, such as~\cite{cohen2018spherical, su2017learning, coors2018spherenet}, and dedicate to solve the distortion issue, there still lacks a suitable dataset to evaluate their approaches on object detection domain.
Both in~\cite{cohen2018spherical, coors2018spherenet}, they experiment the proposed spherical convolutional neural networks on MNIST dataset to perform the classification task.
In~\cite{su2017learning}, they project PASCAL~\cite{Everingham15} to \threesix format and do the object detection.
In addition, in~\cite{coors2018spherenet}, they evaluate the proposed SphereNet on FlyingCars dataset, which is a synthesis dataset combines the real-world background equirectangular images with rendered 3D car models. We evaluate~\cite{coors2018spherenet} the proposed \datasetname dataset in Section~\ref{exp}.

Motivated by the above observation, we present the \datasetname dataset in this paper. \datasetname is the first released and the largest object detection and classification dataset up to now. It consists of $3$k equirectangular indoor images and $90$k Bounding FoVs (BFoVs) annotations among $37$ categories in current version.
\datasetname benchmark is characterized by the following major properties.
\footnotetext{The 80 categories of SUN360 is for image (scene) classification, without instance-level bounding boxes.}
\begin{itemize}
    \item \datasetname is the first released and the largest object detection dataset in \threesix domain, where each image is annotated with $27$ BFoVs on average. The amount of BFoVs provides sufficient data for training and evaluating. 
    \item \datasetname contains the most diverse categories, which include $37$ categories. This will benefit the validation of the generalization capability of any approach.
    %for object detection and classification.
    \item \datasetname is built in real images. It plays an important role since we can easily adapt to real-world applications. Figure~\ref{fig:problem} shows some examples of the images and their annotated BFoVs.
\end{itemize}
For experiments, we benchmark several deep neural network models performing object detection and classification. The best-performing system, FPN, can achieve 33.6\% mean average precision (mAP). In the end, by observing the overall performance of existed methods, we believe \datasetname has not yet saturated and still have room to be improved.

Our contributions in this paper are two-fold:
\begin{itemize}
    \item  We collect the first object detection and classification dataset on \threesix domain which contains $3$k equirectangular indoor images and $90$k BFoVs annotations among $37$ categories.
    \item We comprehensively evaluate three different object detection models on the proposed \datasetname dataset. The results show that standard object detection methods train on the proposed dataset do have large improvements than using NFoVs
\end{itemize}

%%%%%%%%%%%%%%%%%%%%%%%%%%%%%%%%%%%%%%%%%%%%%%%%
%%  Related Work
%%%%%%%%%%%%%%%%%%%%%%%%%%%%%%%%%%%%%%%%%%%%%%%%
\section{Related Work}\label{relate}
\begin{figure*}
\centering
  \includegraphics[width=.95\linewidth]{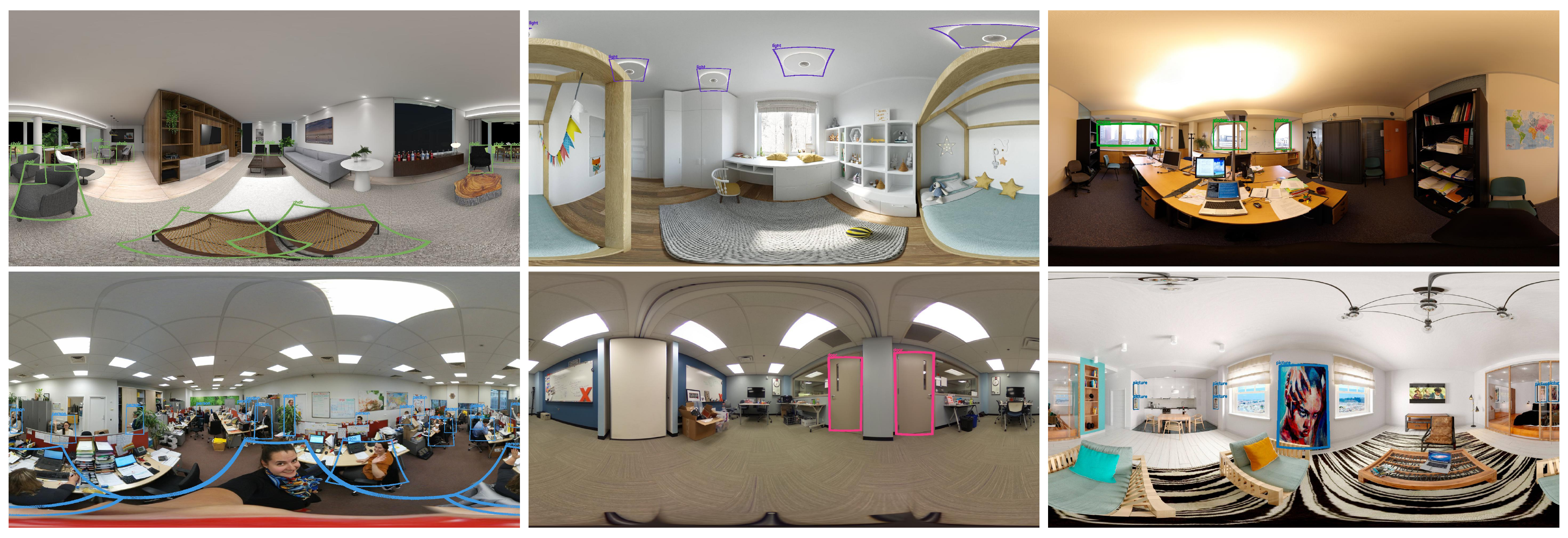}
  \vspace{-1em}
  \caption{Top six categories in \datasetname including chair, light, window (from left to right in the top row), person, door and picture (from left to right in the bottom row). The polygons with different colors indicate different categories. The annotations using BFoVs can better fit the objects in \threesix images. [Best viewed by zooming in.]}
  \label{fig:dataset}
  \vspace{-1em}
\end{figure*}

Datasets in the computer vision research domain play a critical role. They not only provide a means to train and evaluate algorithms, but they also help researchers to explore new and more challenging directions.
Nowadays, the ImageNet dataset~\cite{imagenet_cvpr09} makes breakthroughs in both object classification and detection research. PASCAL, MSCOCO dataset~\cite{Everingham15, lin2014microsoft} with thoroughly annotations provide more complex object recognition tasks to be developed. Recently, with the growing attention on \threesix domain, several \threesix datasets have been proposed. We address this as follows.
\Paragraph{Existed \threesix Dataset.}
\threesix visual is a thriving topic nowadays due to the advance of technology in \threesix cameras.
In recent two years, several datasets in \threesix come up. 
In \cite{su2016activity}, they propose the first \threesix video dataset which contains several outdoor activities.  
Similar to \cite{su2016activity}, Hu et al.~\cite{HuCVPR17} collect a sports~\threesix dataset which covers sports videos and the annotated fixed Normal Field-of-View (NFoV) in each frame.
Furthermore, a highlight detection dataset is proposed in~\cite{yu2018deep}. They crawl the videos from Youtube and Vimeo using keywords `wedding' and `music'. 
In order to match the narrative and content in \threesix videos, Chou et al.~\cite{chou2018self} provides a narrated~\threesix dataset. 
In this dataset, they annotate the objects mentioned in narratives on panoramas chronologically according to the start and end time. 
However, narrated~\threesix dataset only annotates in validation and testing set. 
Furthermore, Wild-$360$~\cite{cheng2018cube} provides the saliency maps to facilitate machine to help users watch the \threesix videos.
SUN360 dataset~\cite{xiao2012recognizing} focuses on scene and viewpoint recognition and has scene/viewpoint labels which are different from the proposed
dataset, \datasetname.
Recently, Yang et al.~\cite{yang2018object} collect a dynamics activities dataset and Coors et al.~\cite{coors2018spherenet} collect a flying cars dataset.
However, the size of the dynamics activities dataset is not large enough. 
Besides, the cars in flying cars dataset are synthesized and added to the images which are hard to apply to real-world directly. 
As a result, we propose \datasetname which is the first release object detection and class classification in \threesix domain. 
We also take the distortion in \threesix images into consideration. The annotation format is tailored for equirectangular images.
Table~\ref{tab:datasecompare} lists the existed \threesix dataset for comparison.

%%%%%%%%%%%%%%%%%%%%%%%%%%%%%%%%%%%%%%%%%%%%%%%%
%%  Dataset
%%%%%%%%%%%%%%%%%%%%%%%%%%%%%%%%%%%%%%%%%%%%%%%%
\vspace{-.5em}
\section{\datasetname Dataset}\label{dataset}
\vspace{-.5em}
\datasetname dataset is collected with $3,335$ indoor images and $89,148$ annotated BFoVs.
In the following, we first introduce the Dataset sections and address each in turn.
In Section~\ref{CategorySelection}, the procedure for object category selection is provided. 
Meanwhile, in Section~\ref{ImageCollection}, we will describe how to collect the candidate images. 
Next, we introduce the novel tool for annotating the \threesix images in Section~\ref{ImageAnnotation}. 
In the end, we provide the statistics about the \threesix Indoor Detection Dataset (\datasetname) in Section~\ref{DatasetStatistics}.
\begin{figure}
\centering
  \includegraphics[width=.9\linewidth]{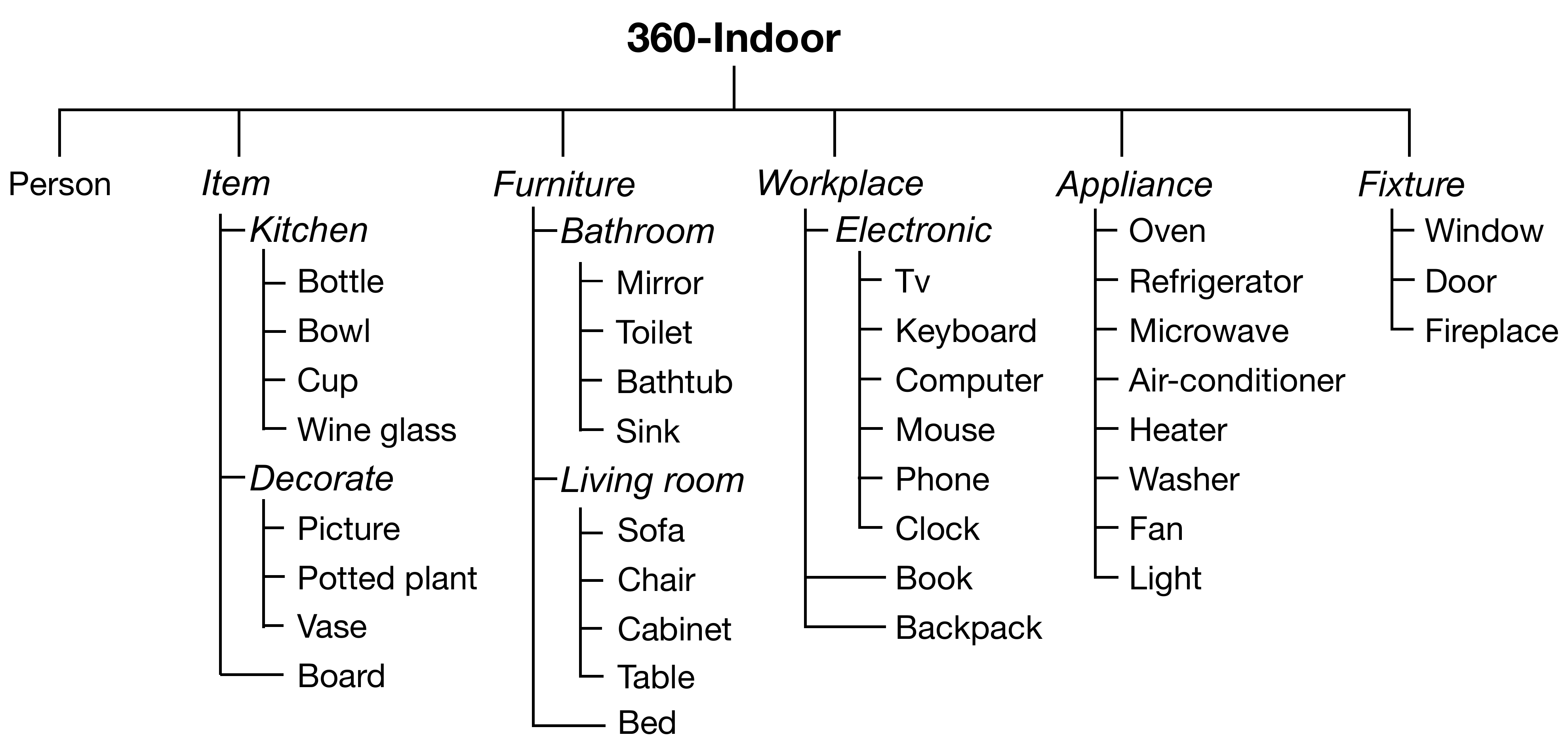}
    \vspace{-.5em}
  \caption{Categories in \datasetname dataset. The italic font denotes the super-categories, and the standardized font denotes the $37$ object categories.}
  \vspace{-1em}
  \label{fig:categoriestree}
\end{figure}
\vspace{-1em}
\subsection{Category Selection}\label{CategorySelection}
\vspace{-0.5em}
For categories selection, we consider the original categories defined in COCO dataset~\cite{lin2014microsoft}. However, since some categories are hard to be seen in the \threesix images, we manually remove these categories having small sizes. For example, we remove tableware such as `spoon' and `fork' as well as the categories in the food super category. In addition, we unify the categories according to the similar properties. For example, we merge `lamp' and `light' into `light'.
Furthermore, we add some categories which are common in the indoor scenes, such as `washer', `heater', `cabinet', etc.
Next, we group the object categories into $5$ super-categories, except for `person'. Each super-category represents a kind of purpose. Since `person' does not belong to any super-categories, we separate it to an independent one.
Figure~\ref{fig:categoriestree} shows the $37$ categories selected for annotation and the super categories in the \datasetname dataset.
\begin{figure}
\centering
\small
  \includegraphics[width=\linewidth]{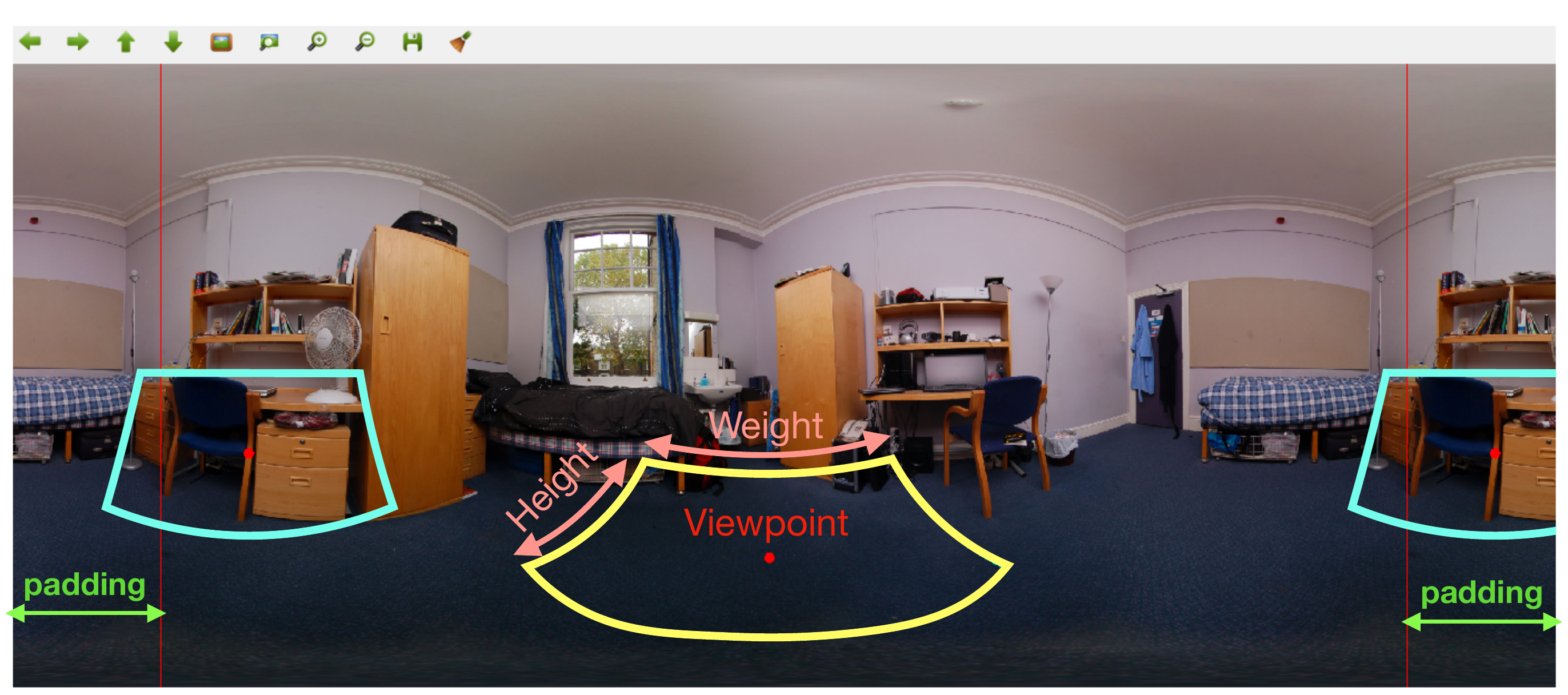}
  \caption{The annotation tool for annotating \datasetname. Each BFoV is represented by the center viewpoint and the width and height of the field-of-view. Annotators first select the category and click the center viewpoint. Then, they use the up/down buttons to enlarge/reduce the height and left/right buttons to enlarge/reduce the width.
  Note that we extend the image boundary using circular padding to take care of objects at image boundary.}
  \label{fig:annotate}
\end{figure}
\vspace{-.5em}
\subsection{Image Collection} \label{ImageCollection}
The images used for \datasetname dataset are collected from Flickr\footnote{\url{https://www.flickr.com/}}, Kuula\footnote{\url{https://kuula.co/}} and Narrated \threesix videos dataset~\cite{chou2018self}.
For images from Flickr, we use related keywords such as `360degrees' and `Equirectangular' to find \threesix images. While Kuula is a \threesix-image sharing platform, so we collect as many images as we can from this website.
However, some of the images retrieved from Flickr and Kuula are duplicated or non-real.
Hence, we first leverage a duplicate finder to remove the duplicated images, and then manually remove the non-real images.
For the Narrated \threesix videos dataset, we take only one frame in each video to avoid redundant scenes.
After collecting the \threesix images from these three resources, we manually select the indoor images and split the images into four different scenes: `activity', `home', `shop', and `work'. 
To balance the categories in the proposed dataset, we remove/add some images in the major/minor scenes. More specifically, scenes ‘work’ and ‘home’ are major in our dataset at first. We remove some images in ‘work’ and ‘home’ scene and add more images in ‘shop’ and ‘activity’.
In the end, we have $3,335$ images in total. All images are with $960\times1,920$ resolution.
\subsection{Image Annotation} \label{ImageAnnotation}
Objects in equirectangular images appear distorted depend on its spatial location, especially in the polar region (the top and bottom region in Figure~\ref{fig:problem}). Therefore, conventional bounding boxes is no longer suitable for labeling \threesix images.
Hence, we choose bounding field of views (BFoVs) presented in~\cite{yang2018object} as annotations in our \datasetname dataset. Unlike the conventional bounding boxes represent by top-left and bottom-right corners ($x_{min}$, $y_{min}$, $x_{max}$, $y_{max}$), BFoVs is defined by ($\phi$, $\theta$, $h$, $w$) (Figure~\ref{fig:annotate}). $\phi$ and $\theta$ are latitude/longitude coordinates of the object’s tangent plane and $h$ is the object height $w$ is the object width.
To facilitate annotators labeling \threesix images, we design an annotation tool which can select the viewpoints and adjust the height and width. The annotation tool is shown in Figure~\ref{fig:annotate}. Annotators are asked to choose the center of the object as the viewpoint (red points in Figure~\ref{fig:annotate}) and using the up/down buttons to adjust the enlarge/reduce height and left/right buttons to enlarge/reduce width. The BFoV will simultaneously show on the image and annotators can easily adjust the BFoVs to match the shape of objects in \threesix images. In addition, since the boundary of the \threesix images continues, the BFoV might across the right and left boundary (i.e., the blue BFoV shown in Figure~\ref{fig:annotate}). For the convenience of the annotators, we pad $45^\circ$ field-of-view region to the left and right side. If the BFoV is across the padding area, the BFoV will show at the other side simultaneously. In order to maintain the unity and coherence of the dataset, each category is labeled by one expert annotator.
\begin{figure*}
\centering
  \includegraphics[width=\linewidth]{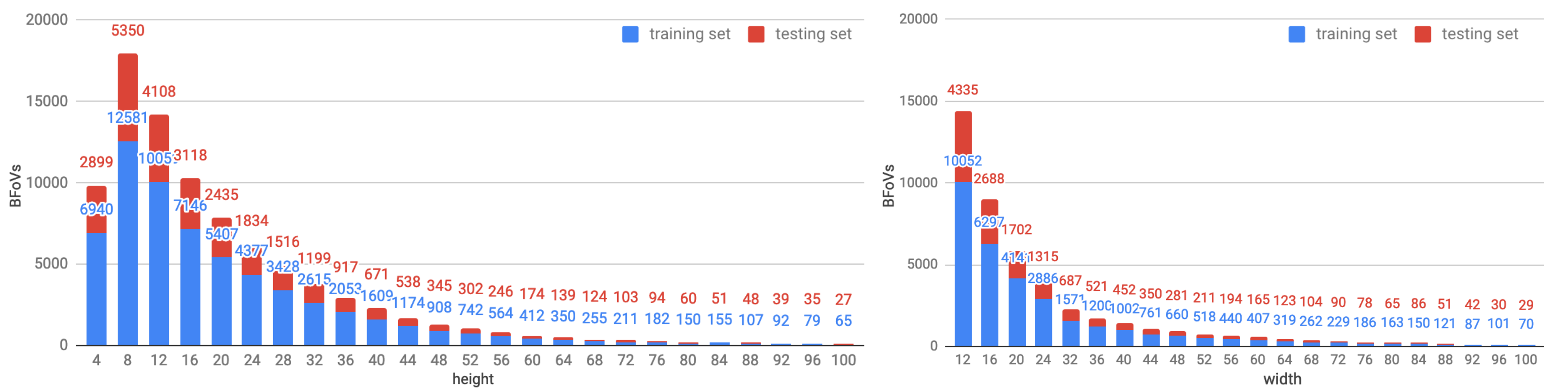}
  \caption{Distribution of height and width in \datasetname.}
  \vspace{-1em}
  \label{fig:wh}
\end{figure*}
\begin{table}
    \small
    \centering
    \caption{Summary of \datasetname dataset.}
    \label{tab:datasetstc}
    \begin{tabular}{c|cc|ccc}
        \toprule
        \textbf{Split}       & \textbf{$\#$images} & \textbf{$\#$BFoV} & \textbf{Avg.} & \textbf{Max.} & \textbf{Min.} \\
        \midrule
        Train       & 2,325 & 62,430 & 26.9 & 211 & 1 \\
        Test       & 1,010 & 26,718 & 26.5 & 223 & 1 \\
        \bottomrule
    \end{tabular}
\end{table}
\subsection{Dataset Statistics} \label{DatasetStatistics}
\begin{figure}
\centering
  \includegraphics[width=.8\linewidth]{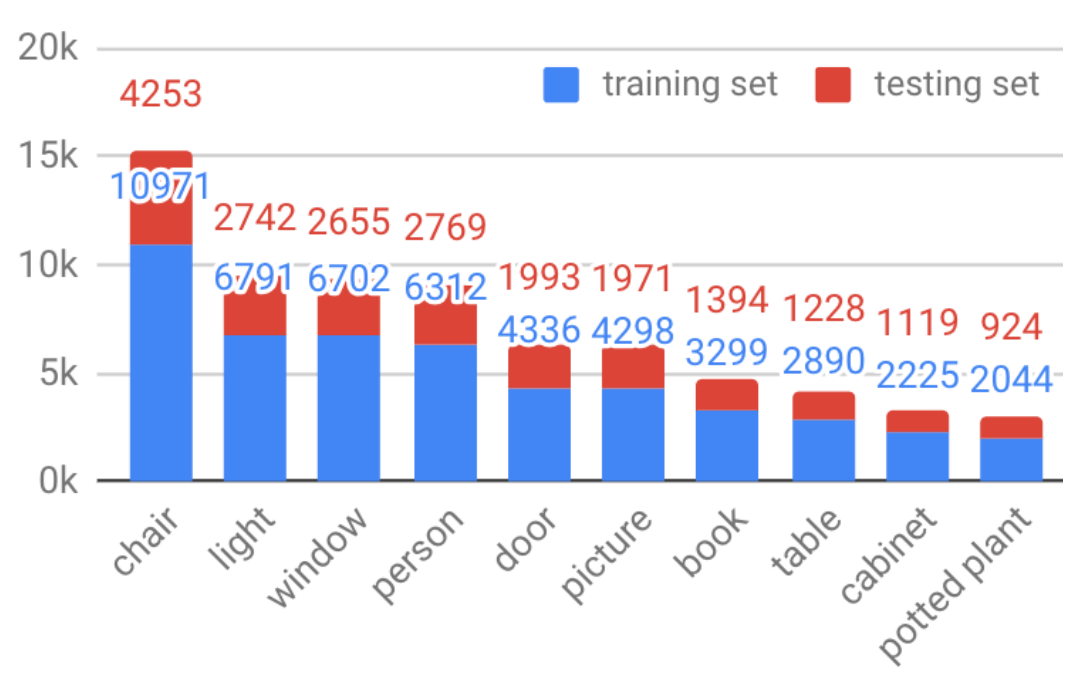}
  \caption{Distribution of top $10$ categories in \datasetname.}
  \vspace{-1em}
  \label{fig:distribution}
\end{figure}

Overall, our \datasetname consists of a total of $3,335$ images. We split the dataset into training/testing set with $70\%$ and $30\%$, respectively. We summarize the statistics of the \datasetname dataset in Table~\ref{tab:datasetstc} and show the distribution of the top $10$ categories in~Figure~\ref{fig:distribution}.
\begin{table}
    \centering
    \small
    \caption{Viewpoints distribution of \datasetname dataset.}
    \label{tab:view}
    \begin{tabular}{c|ccc}
        \toprule
        \textbf{$\#$BFoVs} & 0 to $\pm 30^{\circ}$ & $\pm 30^{\circ}$ to $\pm 60^{\circ}$ & $\pm 60^\circ$ to $\pm 90^\circ$ \\
        \midrule
        Train       & 49,340 & 11,561 & 1,529 \\
        Test       & 21,582 & 4,519 & 617 \\
        \bottomrule
    \end{tabular}
\end{table}
\begin{figure*}
\centering
  \includegraphics[width=.9\linewidth]{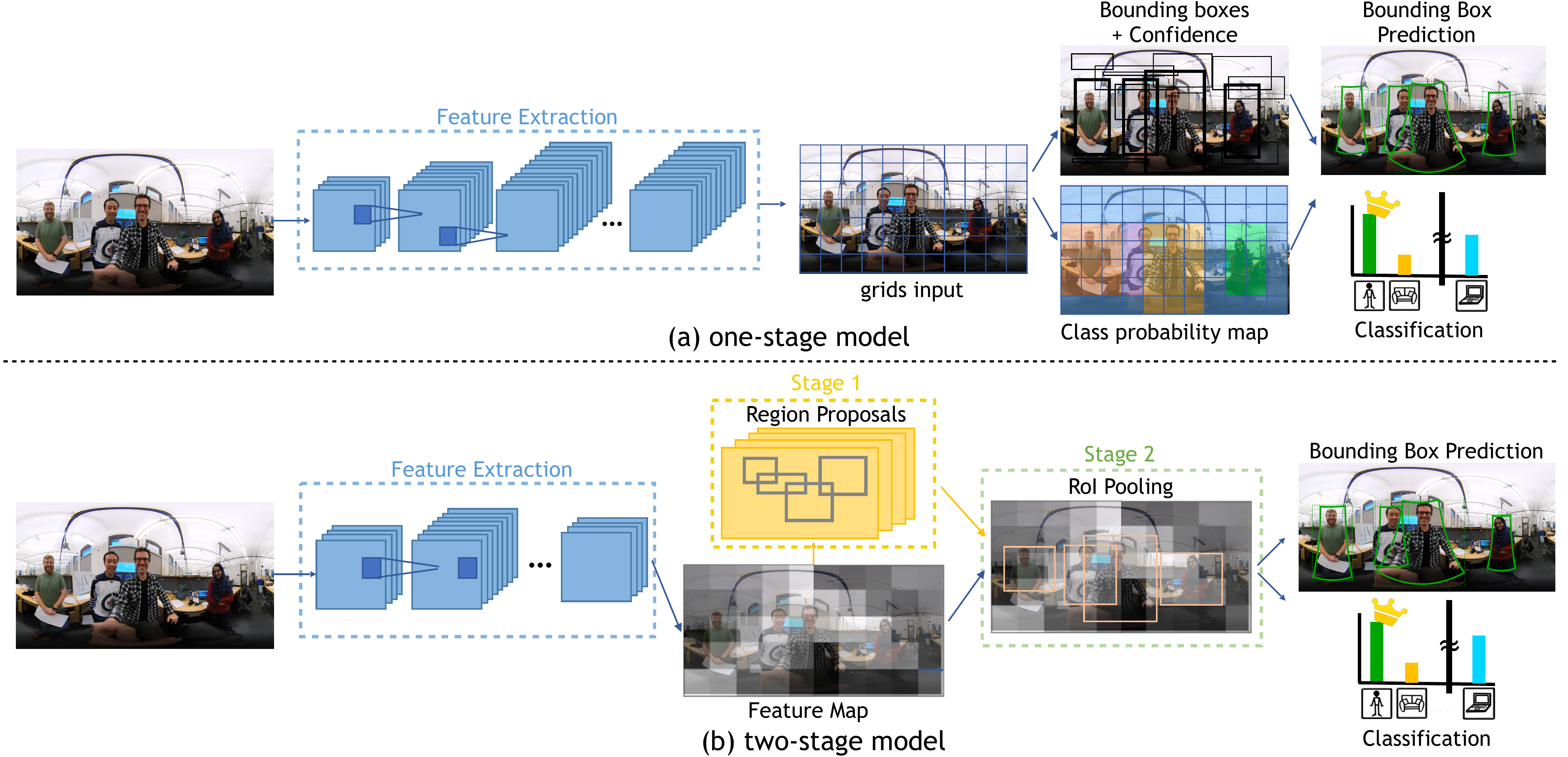}
  \caption{Approach. Summarization of two different object detection approaches. (a) shows the framework of one stage detection. The model utilizes CNNs to extract the visual representation for objects and directly predict bounding box information, foreground/background confidence and class confidence. (b) shows the framework of two-stage detection. After deriving feature map, the model predicts the bounding box with foreground confidence score at first. Then, using a detector to aggregate feature map and proposed regions, the model is able to predict bounding boxes and give class probabilities.}
  \vspace{-1em}
  \label{fig:approach}
\end{figure*}
In addition, we also provide the statistics of the distribution of viewpoints, height, and width of \datasetname dataset. The results are shown in Table~\ref{tab:view} and Figure~\ref{fig:wh}, respectively. 
For viewpoints distribution, we first separate latitudes into $\phi \in \Phi = \{0, \pm30, \pm60, \pm90\}$ according to the distortion of objects. Between 0 to $\pm$30$^{\circ}$, object distortion is less. The distortion will become severe when $\phi$ become bigger. 
In latitude from $\pm$60$^{\circ}$ to $\pm$90$^{\circ}$, objects will have the largest distortion which has the most significant difference from conventional images. 
As illustrated in Figure~\ref{tab:view}, most of the objects appear in latitudes from $0$ to $\pm$30$^{\circ}$. This is a common scenario since objects appear in the middle of the indoor scene.
In addition, in high latitudes region ($\pm$60$^{\circ}$ to $\pm$90$^{\circ}$), \datasetname dataset has sufficient portion objects which are able to help machine to recognize. 
For height and width distribution, they are alike in training and testing set. The mode of height and width in training and testing set are both $8$ angular dimensions.
Please note that the height and width do not correspond to the height and width in conventional bounding boxes since the regions in the equirectangular projection are not rectangular. For example, in the polar region, although the height and width are small, the projected region in the equirectangular image will cover a large portion.

%%%%%%%%%%%%%%%%%%%%%%%%%%%%%%%%%%%%%%%%%%%%%%%%
%%  Approach
%%%%%%%%%%%%%%%%%%%%%%%%%%%%%%%%%%%%%%%%%%%%%%%%
\section{Approaches to Object Detection}\label{sec:approach}
We briefly describe different object detection approaches that we benchmark on our proposed \datasetname dataset. 
Most of the state-of-the-art approaches for object detection are based on the Convolutional Neural Networks (CNNs) which can capture spatial correlation by the filter with patterns. As this dimension of research achieves better performance than the traditional hand-craft feature (e.g., the histogram of oriented gradients, Bag-of-visual Words model), we summarize CNN-based approaches as two kinds of methods, one-stage object detection, and two-stage object detection. 
In one-stage object detection approach~\cite{wei_2016_SSD,redmon2016you,joseph_2017_yolo9000,redmon2018yolov3} (as shown in Figure~\ref{fig:approach} (a)), given an input image, the model utilizes CNNs to extract the visual representation for objects and directly predict bounding box information, foreground/background confidence and class confidence in every grid cell. 
The model first divides the image feature into grids. Each grid cell predicts $B$ bounding boxes and confidence scores for those boxes. The confidence scores are provided based on how confident the model regards the box contains an object and also how accurate.
Each grid cell also predicts $C$ conditional class probabilities. These probabilities are conditioned on the grid cell containing an object.
In testing phase, class-specific confidence scores can be derived by multiplying the conditional class probabilities and the individual box confidence predictions. These scores encode both the probability of that class appearing in the box and how well the predicted box fits the object.
In sum, the one-stage model will optimize the localization task and classification task at the same time by framing object detection as a single regression problem, straight from image pixels to bounding box coordinates and class probabilities.

On the other hand, in the two-stage object detection approach~\cite{girshick2014rich, uijlings2013selective, girshick2015fast, dai2016instance, he2017mask, ren2015faster, dai2016r, lin2017feature}, given an input image, the model will use CNNs to extract feature and predict the bounding box with foreground confidence score at first. Then, using a detector to aggregate feature map and proposed regions, the model is able to predict bounding boxes and give class probabilities (as shown in Figure~\ref{fig:approach} (b)).
The model first uses a deep fully convolutional network to propose regions in stage one (the yellow part in Figure~\ref{fig:approach} (b)). To generate region proposals, sliding windows are used over the feature map. 
At each sliding-window location, an anchor is centered and is associated with a scale and aspect ratio. Leveraging non-maximum suppression, candidate region proposals can be derived.
In stage two, a region-based CNN (R-CNN) detector uses the proposed regions (the green part in Figure~\ref{fig:approach} (b)). The detector takes the feature map and proposed regions as inputs to predict bounding boxes and give class probabilities. The entire system is a single, unified network for object detection. Using the terminology of neural networks with attention mechanisms, the region proposal module tells the R-CNN module where to look.
There are multiple ways to get more general features. The first one is conventional extract feature~\cite{ren2015faster}. The second one is using multi-level semantic information from different layers~\cite{lin2017feature}.

Among all CNN-based state-of-the-art methods for object detection, we mainly investigate and evaluate in terms of three directions on our \datasetname dataset: one one-stage object detection, YOLOv3~\cite{redmon2018yolov3}, and two two-stage object detection, a conventional feature extraction, Faster R-CNN~\cite{ren2015faster} and multi-level feature extraction, FPN~\cite{lin2017feature}. 

In addition, there are several spherical CNNs dedicate to eliminate the distortion problem in the equirectangular images, such as~\cite{su2017learning, boomsma2017spherical, cohen2018spherical, coors2018spherenet, esteves2018learning}. We choose SphereNet~\cite{coors2018spherenet} as its ease of integration into an object detection model. In one-stage and two-stage object detection, we replace the \texttt{Conv2d} and \texttt{MaxPool2d} in the feature extraction to \texttt{SphereConv2d} and \texttt{SphereMaxPool2d}~\cite{coors2018spherenet}. The rest of the object detection model remains the same.

%%%%%%%%%%%%%%%%%%%%%%%%%%%%%%%%%%%%%%%%%%%%%%%%
%%  Experiment
%%%%%%%%%%%%%%%%%%%%%%%%%%%%%%%%%%%%%%%%%%%%%%%%
\section{Experiments}\label{exp}
\begin{table*}
\footnotesize
\centering
    \caption{Experiment on \datasetname dataset. We compare with the baseline which is pretrained on COCO as well as different backbone networks. mAP (overlap) denotes we only use the overlap categories between COCO and \datasetname to calculate mAP. mAP denotes using the result from all categories in \datasetname dataset.}
    \label{tab:exp}
    \begin{tabular}{c|c|c|c|c}
    \toprule
    \bf{Model} & \bf{Backbone} & \bf{Anchors} & \bf{mAP (overlap) (\%)} & \bf{mAP (\%)}\\
    \midrule
    \multirow{3}{*}{YOLOv3} & baseline (DarkNet53 trained on COCO) & kmeans++ for COCO (9)~\cite{redmon2018yolov3} & 10.9 & - \\ 
     & ResNet50 trained on \datasetname &kmeans++ for COCO (9)~\cite{redmon2018yolov3} & 11.9 & 12.4 \\ 
     & DarkNet53 trained on \datasetname & kmeans++ for COCO (9)~\cite{redmon2018yolov3} & \bf{22.7} & \bf{24.5} \\ 
     % & DarkNet53 &  &  & 27.2 \\ 
     \midrule
    \multirow{3}{*}{Faster R-CNN} & baseline (ResNet101 trained on COCO) & default~\cite{ren2015faster} & 11.5 & - \\
     & ResNet50 trained on \datasetname & default~\cite{ren2015faster} & 24.5 & 29.1 \\ 
     & ResNet101 trained on \datasetname & default~\cite{ren2015faster} & \bf{25.3} & \bf{30.2} \\ 
     % & ResNet101 &  &  & 30.5 \\ 
     \midrule
    \multirow{3}{*}{FPN} & baseline (ResNet101 trained on COCO) & default~\cite{lin2017feature} & 12.2 & - \\
     & ResNet50 trained on \datasetname& default~\cite{lin2017feature} & 27.4 & 33.1 \\
     & ResNet101 trained on \datasetname& default~\cite{lin2017feature} & \bf{28.2} & \bf{33.6} \\
     % & ResNet101 &  &  & \\
     \bottomrule
     \end{tabular}
\end{table*}
We conduct three widely used object detection approaches on \datasetname dataset. The approach can be separated into two types: (1) one-stage object detection approach (e.g., YOLOv3~\cite{redmon2018yolov3}) and (2) two-stage object detection approach (e.g., Faster R-CNN~\cite{ren2015faster} and FPN~\cite{lin2017feature}).
In addition, the conventional CNNs also replace by the SphereNet in order to learn the invariance of these distortions.
\vspace{-1em}
\paragraph{Bounding Field-of-View Transform}
To match the input formats of the above-mentioned approaches, we first transform the bounding FoVs to conventional bounding boxes.
The vertexes of the conventional bounding boxes can be derived by projecting ground truth annotations to the tangent plane. That is, we use a mapping function from~\cite{chou2018self} to map the annotations in spatial coordinates to tangent coordinates. The mapping function takes viewpoints, width, and height as input, and outputs the corresponding pixels in tangent coordinates. Hence, the projected pixels can be derived. 
After having the projected pixels, we choose the boundary of these pixels to draw the transformed bounding boxes.
Therefore, the $(x_{min}, y_{min}, x_{max}, y_{max})$ in the conventional bounding box can be derived and fed into the object detection networks.
\vspace{-1em}
\paragraph{Faster R-CNN.} We use the official settings to implement Faster R-CNN.
The anchor scales are set as $\{64^2, 128^2, 256^2, 512^2\}$ with aspect ratio $\{\frac{1}{2}, 1, 2\}$ which are the same as the setting in COCO dataset.
We use ResNet101~\cite{he2016deep} pretrained on ImageNet as our backbone as it gives us better mAP. 
We train Faster-RCNN with $1$ GPU with batch size $1$, $12,000$ RoIs per image before applying NMS to RPN proposals and $2000$ after applying NMS to RPN proposals. 
We use SGD with momentum $0.9$ and weight decay $10^{-4}$. The learning rate is set to $0.001$.
\vspace{-1em}
\paragraph{FPN.}
\label{sec::Experiment::para::fpn}
We follow the official implementation. The anchor scales are set as $\{32^2, 64^2, 128^2, 256^2, 512^2\}$ with aspect ratio $\{\frac{1}{2}, 1, 2\}$.
We also use the same method to map RoI to pyramid level. 
Instead of using RoI pooling, we use RoI align~\cite{he2017mask} to extract the proposed candidates for second-stage. 
We use ResNet101~\cite{he2016deep} pretrained on ImageNet as our backbone as it gives us better mAP. 
We train FPN with $1$ GPU with batch size $2$, $512$ RoIs per image to train first-stage and $2,000$ RoIs per FPN level to train second-stage. 
We use SGD with momentum $0.9$ and weight decay $10^{-4}$. The learning rate is set to $0.0025$ because our batch size is $8$ time smaller than official. 
The running statistic of BatchNorm is fixed following standard practice for small batch size. 
The images' resolution is all resized to $960 \times 960$ as it shows good accuracy and computation resource trade-off.
\vspace{-1em}
\paragraph{YOLOv3.}\label{sec::Experiment::para::yolo_v3}
We leverage the following training strategy proposed by the J. Redmon \etal~\cite{redmon2018yolov3}: multi-scale training, data augmentation and batch normalization. 
We choose the Darknet-53 pretrained on ImageNet as our backbone, which performance approaches the ResNet-101 and 1.5$\times$ faster (as mentioned in~\cite{redmon2018yolov3}). 
However, we notice that the original preprocess of YOLOv3 is not suitable for the high respect ratio image. Because images will be padded to a square shape and resize to $416 \times 416$, the final predictions will have low precision due to lack of feature.
Hence, we directly resize \threesix images with the original size $960\times1,920$ to $960\times960$.
We use SGD with momentum $0.9$ and weight decay $10^{-3}$.
The learning rate is set as $5*10^{-4}$ which is the same as in~\cite{redmon2018yolov3}.
\vspace{-1em}
\paragraph{Baseline.} To demonstrate the effectiveness of the proposed \datasetname dataset, we further consider the following models. We take the COCO dataset pretrained model and directly use the testing set of \datasetname dataset to evaluate. For every approach, we use ResNet101~\cite{he2016deep} as the backbone network and use the same anchor box settings from their reference.
\subsection{Discussion}
\begin{figure*}
\centering
  \includegraphics[width=.95\linewidth]{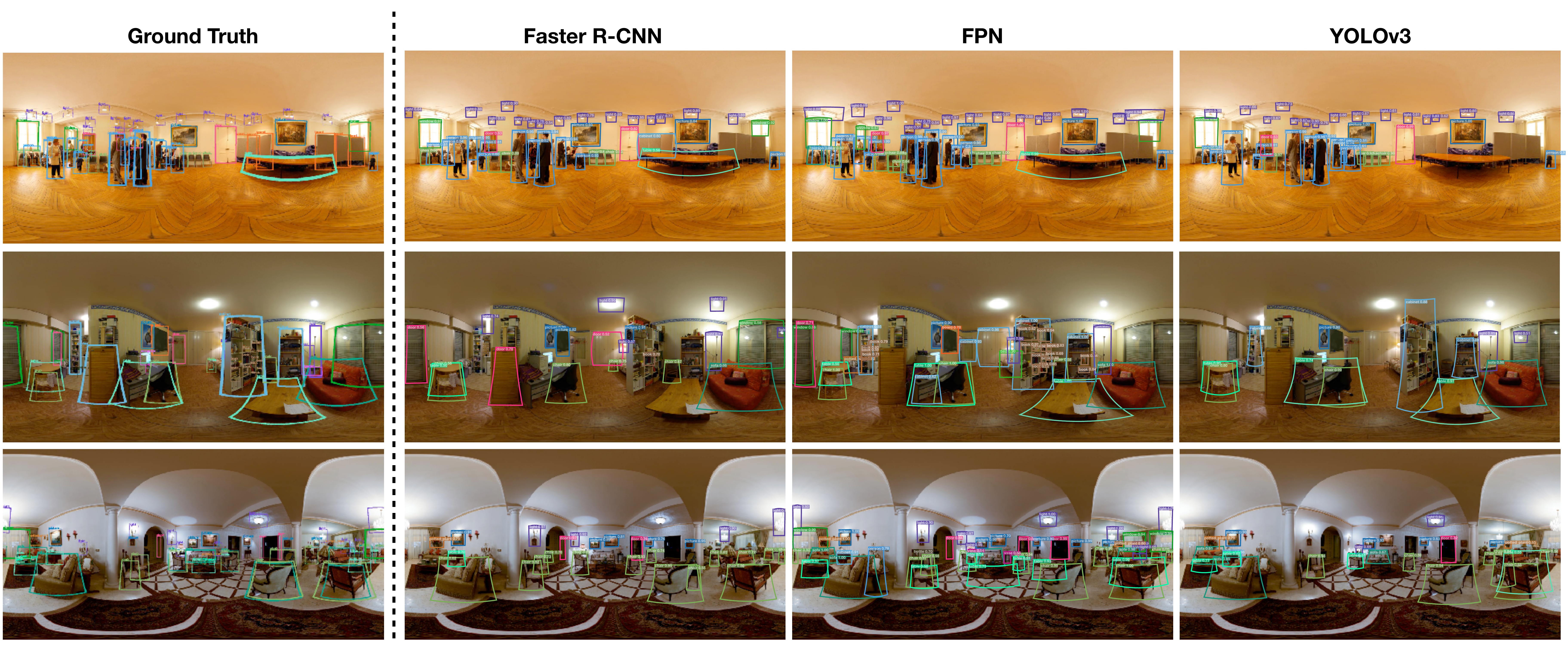}
  \caption{Qualitative results. We show three examples to illustrate the results of object detection from different approaches and ground truth. FPN is able to detect the objects with distortion and small objects. Hence, it achieves the best performance among the three approaches.}
  \vspace{-1em}
  \label{fig:qualitative}
\end{figure*}
\begin{table}
    \footnotesize
    \centering
    \caption{Analysis of anchor proposals in YOLOv3. We show the comparison of different anchor proposal setting.}
    \label{tab:yoloanchor}
    \begin{tabular}{c|c|c}
        \toprule
        \textbf{Procedure} & \textbf{Anchor Proposals} &\textbf{mAP (\%)} \\
        \midrule
        \multirow{3}{*}{COCO} &  (10$\times$13),(16$\times$30), (33$\times$23), &\multirow{3}{*}{24.5}\\
         & (30$\times$61),(62$\times$45), (59$\times$119),&\\
         & (116$\times$90),(156$\times$198), (373$\times$326)&\\
         \midrule
         \multirow{3}{*}{2.3$\times$} & (23$\times$30),(37$\times$69), (76$\times$53), &\multirow{3}{*}{22.4}\\
         & (69$\times$141),(143$\times$104), (136$\times$275),&\\
         & (268$\times$208),(358$\times$455), (858$\times$750)&\\
         \midrule
        \multirow{3}{*}{\datasetname} & (17$\times$21),(11$\times$60), (31$\times$58), &\multirow{3}{*}{\textbf{27.2}}\\
         & (38$\times$128),(101$\times$113), (65$\times$242),&\\
         & (165$\times$249),(327$\times$313), (959$\times$201)&\\
        \bottomrule
    \end{tabular}
    \vspace{-1em}
\end{table}
The results are shown in Table~\ref{tab:exp} 
indicate that all three detectors trained on our \datasetname dataset significantly outperform detectors trained on COCO dataset. We utilize mean average precision (mAP) as the evaluation method.
We first use the pretrained detectors and directly test on the proposed \datasetname testing set, which refers to the first row in each model (baseline). 
In addition, mAP (Overlap) denotes that we only use the overlap categories between COCO and \datasetname to calculate mAP.
Comparing with the detectors fine-tune on \datasetname dataset (third row in each model), it is critical to train detectors on our \datasetname dataset to achieve high object detection accuracy on equirectangular images.
Compare with three approaches, FPN achieves the best performance which indicates that it has a better ability to deal with the distorted images.
Among the three detectors, YOLOv3 achieves slightly worse accuracy. Since YOLOv3 is sensitive to the anchor proposals, we further conduct an analysis of the anchor proposals.
\vspace{-1em}
\paragraph{Analysis of anchor proposals in YOLOv3.}
We evaluate three kinds of anchor proposals with YOLOv3. Firstly, we use anchor boxes size calculated from COCO datasets, which is the original setting in YOLOv3. 
Secondly, we modify the anchor boxes size to be $2.3\times$ bigger than the anchor boxes calculated from COCO datasets since our input image size is $2.3\times$ large than the original setting.
Finally, we directly calculate $9$ anchor boxes size from \datasetname dataset using kmeans++ as suggested in~\cite{redmon2018yolov3}.

The results in Table~\ref{tab:yoloanchor} show that it is critical to calculating anchor boxes size from our \datasetname dataset. We also found that YOLOv3 takes longer to converge during training when anchors boxes are not calculated from our \datasetname dataset. Our \datasetname dataset is again the key contributor to the performance improvement for YOLOv3.
\vspace{-1em}
\paragraph{Object detection model with SphereNet.}
To better encodes invariance against such distortions explicitly into convolutional neural networks, we replace the \texttt{Conv2d} and \texttt{MaxPool2d} in the conventional CNNs to \texttt{SphereConv2d} and \texttt{SphereMaxPool2d}, respectively. The results are shown in Table~\ref{tab:sphere}. 
Comparing with row one and three, \texttt{Conv2d} (pretrained on ImageNet) and \texttt{MaxPool2d} is better than \texttt{SphereConv2d}/\texttt{SphereMaxPool2d}.
We think the reason is that \texttt{SphereConv2d} is trained from scratch, the model needs more time to converge. This indicates the importance of pretrained model.
For a fair comparison, we train \texttt{Conv2d} from scratch (second row in Table~\ref{tab:sphere}) and use \texttt{MaxPool2d}. 
We can notice that \texttt{Conv2d} (w/o pretrain)/\texttt{MaxPool2d} is slightly better than \texttt{SphereConv2d}/\texttt{SphereMaxPool2d}.
We argue that because conventional object detectors are designed for the normal images, the compatibility of the conventional CNNs and object detectors is better than SphereNet.
Hence, the object detectors specialize for \threesix images is needed. As a result, we believe the proposed \datasetname dataset provides a good start point for \threesix domain in future studies. \datasetname is large enough to train detectors and benefit validation of the generalization capability of any approach.
\vspace{-2em}
\paragraph{Qualitative results.}
The qualitative results are shown in Figure~\ref{fig:qualitative}. To better compare with the ground truth, we project rectangular boxes to spherical plane. As illustrated in Figure~\ref{fig:qualitative}, FPN can detect the objects more correctly, and even the objects are with more distortion or small.
\vspace{-.5em}

\begin{table}
    \footnotesize
    \centering
    \caption{Performance of object detection models with and without SphereNet. We show the comparison with different model settings.}
    \label{tab:sphere}
    \begin{tabular}{c|c|c}
        \toprule
        \textbf{Procedure} & \textbf{Settings} & \textbf{mAP (\%)} \\
        \midrule
        Faster R-CNN & \texttt{Conv2d} (pretrain)/ \texttt{MaxPool2d} & \textbf{30.2}  \\
        Faster R-CNN &  \texttt{Conv2d} (w/o pretrain)/\texttt{MaxPool2d} & 24.1 \\
        Faster R-CNN & \texttt{SphereConv2d}/\texttt{SphereMaxPool2d} & 21.7 \\
        \bottomrule
    \end{tabular}
    \vspace{-1em}
\end{table}
\vspace{-1em}

%%%%%%%%%%%%%%%%%%%%%%%%%%%%%%%%%%%%%%%%%%%%%%%%
%%  Conclusion
%%%%%%%%%%%%%%%%%%%%%%%%%%%%%%%%%%%%%%%%%%%%%%%%
\section{Conclusion and Future Work}\label{conclude}

We present a real-world \threesix panoramic object detection dataset, \datasetname, which is a new benchmark for visual object detection and class recognition in \threesix images.
It consists of complex indoor images containing common objects and the intensive annotated bounding field-of-view. To the best of our knowledge, it is the largest one in category numbers and number of bounding boxes.
Extensive experiments on the detection methods show that training using our \datasetname dataset is the key to achieve state-of-the-art accuracy on 360$^\circ$ images.
Thus, our \datasetname dataset can contribute to future development on applications (e.g., robot perception and virtual reality) requiring detecting objects in 360$^\circ$ images.
In the future, we aim at developing dedicated object detectors overcoming image distortion and leveraging context from the complete field-of-view in a scene.

\paragraph{Acknowledgments.}
We thank MOST Joint Research Center for AI Technology and All Vista Healthcare for their supports.
%We thank XXX and XXX for helpful comments and discussion.

{\small
\bibliographystyle{ieee}
\bibliography{egbib}
}

\end{document}